\documentclass[10pt,twocolumn,letterpaper]{article}

\usepackage{iccv}
\usepackage{times}
\usepackage{epsfig}
\usepackage{graphicx}
\usepackage{amsmath}
\usepackage{amssymb}
\usepackage{algpseudocode}

\usepackage[ruled,linesnumbered]{algorithm2e}
\usepackage{multirow}
\usepackage{array}
\usepackage{bbding}
\usepackage{latexsym}
\usepackage{balance}

\usepackage[pagebackref=true,breaklinks=true,letterpaper=true,colorlinks,bookmarks=false]{hyperref}

\iccvfinalcopy % *** Uncomment this line for the final submission

\ificcvfinal\fi

\begin{document}

%%%%%%%%% TITLE
\title{IDM: An Intermediate Domain Module for Domain Adaptive Person Re-ID}

\author{{Yongxing Dai{$^{1}$}} \quad Jun Liu{$^{2}$} \quad Yifan Sun{$^{3}$} \quad Zekun Tong{$^{4}$} \quad Chi Zhang{$^{3}$} \quad  Ling-Yu Duan{$^{1, 5}$}\thanks{Corresponding Author.} \\
	\small
	$^{1}$\	Institute of Digital Media (IDM), Peking University, Beijing, China ~~ $^{2}$\, Singapore University of Technology and Design, Singapore \\ 
	\small $^{3}$\ Megvii Technology ~~ $^{4}$\ National University of Singapore, Singapore ~~  ${^5}$\ Peng Cheng Laboratory, Shenzhen, China \\
	\small
	{\tt\small \{yongxingdai, lingyu\}@pku.edu.cn, jun\_liu@sutd.edu.sg, sunyf15@tsinghua.org.cn}
	}

\maketitle
\ificcvfinal\thispagestyle{empty}\fi

%%%%%%%%% ABSTRACT
\begin{abstract}
Unsupervised domain adaptive person re-identification (UDA re-ID) aims at transferring the labeled source domain's knowledge to improve the model's discriminability on the unlabeled target domain. From a novel perspective, we argue that the bridging between the source and target domains can be utilized to tackle the UDA re-ID task, and we focus on explicitly modeling appropriate intermediate domains to characterize this bridging. Specifically, we propose an Intermediate Domain Module (IDM) to generate intermediate domains' representations on-the-fly by mixing the source and target domains' hidden representations using two domain factors. Based on the ``shortest geodesic path'' definition, \textit{i.e.,} the intermediate domains along the shortest geodesic path between the two extreme domains can play a better bridging role, we propose two properties that these intermediate domains should satisfy. To ensure these two properties to better characterize appropriate intermediate domains, we enforce the bridge losses on intermediate domains' prediction space and feature space, and enforce a diversity loss on the two domain factors. The bridge losses aim at guiding the distribution of appropriate intermediate domains to keep the right distance to the source and target domains. The diversity loss serves as a regularization to prevent the generated intermediate domains from being over-fitting to either of the source and target domains. Our proposed method outperforms the state-of-the-arts by a large margin in all the common UDA re-ID tasks, and the mAP gain is up to 7.7\% on the challenging MSMT17 benchmark. Code is available at \url{https://github.com/SikaStar/IDM}.
\end{abstract}

\begin{figure}[htp]
\begin{center}
\includegraphics[width=0.8\linewidth]{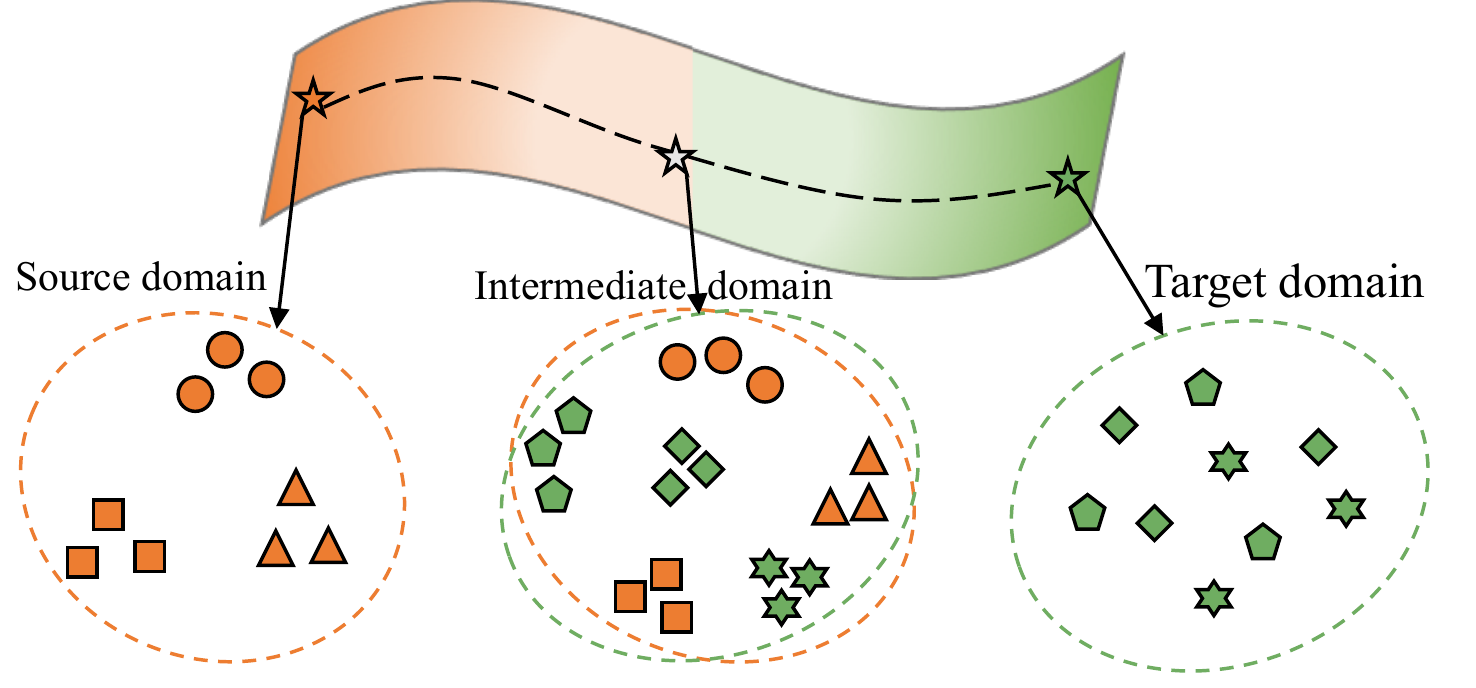}
\end{center}
\caption{Illustration of our main idea. Assuming that the source and target domains are located in a manifold, there can be some intermediate domains along with the path between the two extreme domains. By utilizing appropriate intermediate domains to bridge the source and target, the source knowledge can be better transferred to the target domain. Different colors and shapes represent different domains and different person identities respectively.}
\label{fig:intro}
\vspace{-2em}
\end{figure}

%%%%%%%%% BODY TEXT
\section{Introduction}
Person re-identification (re-ID) aims to identify the person across the non-overlapped cameras. Though fully supervised re-ID methods \cite{sun2018beyond,wang2018learning,luo2019strong,zhang2020relation} have achieved great progress in recent years, it takes much time and effort to label the person images and thus can be difficult to be well applied to some practical scenarios. To overcome this problem, researchers have been focusing on studying unsupervised domain adaptive (UDA) re-ID \cite{wei2018person,fu2019self,song2020unsupervised}, whose goal is to transfer the knowledge from the labeled source domain to improve the model's discriminability on the unlabeled target domain. UDA re-ID is a challenging problem because the source and target domains can have two extreme distributions, and there can be no overlap between the two domains' label space. In this paper, we are devoted to studying the problem of UDA re-ID.

Though many existing methods \cite{wei2018person,fu2019self,ge2020self} have made great progress in UDA re-ID, they have not explicitly considered the bridging between the two extreme domains, \textit{i.e.,} what information about the similarity/dissimilarity of the two domains can be utilized to tackle the UDA re-ID task. For example, GAN transferring methods \cite{wei2018person,deng2018image} use GANs to translate images' style across domains and then focus on a single domain's supervised feature learning. Fine-tuning methods~\cite{song2020unsupervised,fu2019self,dai2020dual} only use the source domain to obtain the pretrained model and then focus on the feature learning in the target domain. Joint training methods~\cite{zhong2020learning,ge2020self} just combine the source and target data together for training. From a novel perspective, we explicitly and comprehensively investigate into the bridging between the source and target domains, and argue that using this bridging can help better adapt between two extreme domains in UDA~re-ID.

In a spirit of the golden mean, we propose to use the deep model to learn intermediate domains which can bridge the source and target domains and thus ease the UDA re-ID task. The source domain's knowledge is hard to be directly transferred to the target domain because there can be huge shift between the two extreme domains' distributions. As shown in Figure~\ref{fig:intro}, assuming the source and target domains are located in a manifold, an appropriate ``path'' may exist to bridge the two extreme domains along which the source domain's knowledge can be smoothly transferred to guide the learning of the target domain. Specifically, the appropriate intermediate domains should be located along with this ``path'' to help the gradual adaptation between the two extreme domains. For example, if an intermediate domain is closer to the source domain, the source reliable labels can be more leveraged. On the contrary, the target domain's intrinsic distribution can be more exploited.

To model the characteristics of appropriate intermediate domains, we propose a plug-and-play intermediate domain module (IDM).
The IDM module can be plugged at any hidden stage of a network, which will generate intermediate domains' representations on-the-fly to gradually bridge the source and target domains in a joint training framework. 
Specifically, the IDM module takes some hidden stage's representations of the source and target domains as the input, and obtains two domain relevance variables that we call domain factors. We use these two domain factors to mix the hidden stage's representations of both source and target domains, where the mixed representations can represent the characteristics of intermediate domains. All these representations from the source, target and intermediate domains are fed into the network's next stage on-the-fly.

However, there may be infinite intermediate domains when mixing the source and target domains, and only a part of these domains may be more helpful to smoothly adapt between the two extreme domains (the source and target domains). Inspired by the traditional works \cite{gong2012geodesic,gopalan2013unsupervised} that construct intermediate domains using kernel-based methods in a manifold, we first propose the ``shortest geodesic path'' definition that the appropriate intermediate domains should lie on the shortest geodesic path connecting the source and target domains. From this definition, we further propose two properties where appropriate intermediate domains should satisfy: (1) Keep the right distance to the source and target domains. (2) Be diverse enough to balance the source and target domains' learning and avoid being over-fitting to either of them. 
For the first property above, we enforce the bridge losses on both feature and prediction spaces of intermediate domains.
For the second property, we propose a diversity loss by maximizing the standard deviation of the domain factors generated by the IDM module. 
By enforcing these losses on the learning of the IDM module, we can model the characteristics of appropriate intermediate domains to better transfer the source knowledge to improve the model's discriminability on the target domain.

Our contributions can be summarized as follows.
(1)~To the best of our knowledge, we are the first to explicitly consider how to utilize the characteristics of intermediate domains as the bridge to better transfer the source knowledge to the target domain in UDA re-ID.
Specifically, we propose a plug-and-paly IDM module to generate intermediate domains on-the-fly, which will smoothly bridge the source and target domains to better adapt between the two extremes to ease the UDA re-ID task.
(2)~To make the IDM module learn more appropriate intermediate domains, we propose two properties of these domains, and design the bridge losses and diversity loss to satisfy these properties.
(3)~Our method outperforms state-of-the-arts by a large margin on all the common UDA re-ID tasks.

\section{Related Work}

\textbf{Unsupervised Domain Adaptation.} 
A line of works convert the UDA into an adversarial learning task \cite{ganin2015unsupervised,tzeng2017adversarial,long2018conditional,russo2018source}. 
Another line of works use various metrics to measure and minimize the domain discrepancy, such as MMD \cite{long2015learning} or other metrics \cite{sun2016deep,zhuo2017deep,long2017deep,kang2018deep}.
Another line of traditional works \cite{gong2012geodesic,cui2014flowing,gopalan2013unsupervised} try to bridge the source and target domains based on intermediate domains. In traditional methods \cite{gong2012geodesic,gopalan2013unsupervised}, they embed the source and target data into a Grassmann manifold and learn a specific geodesic path between the two domains to bridge the source and target domains, but they are not easily applied to the deep models. In deep methods \cite{gong2019dlow,cui2020gradually}, they either use GANs to generate a domain flow by reconstructing input images on pixel level\cite{gong2019dlow} or learn better domain-invariant features by bridging the learning of the generator and discriminator \cite{cui2020gradually}. 
However, reconstructing images may not guarantee the high-level domain characteristics in the feature space, and learning domain-invariant features may be not suitable for two extreme domains in UDA re-ID.
Different from the above methods, we propose a specific IDM module to model the intermediate domains, which can be easily inserted into the existing deep networks. Instead of hard training for GANs or reconstructing images, our IDM module can be learned in an efficient joint training scheme.
\begin{figure*}[htp]
\begin{center}
\includegraphics[width=1.\linewidth]{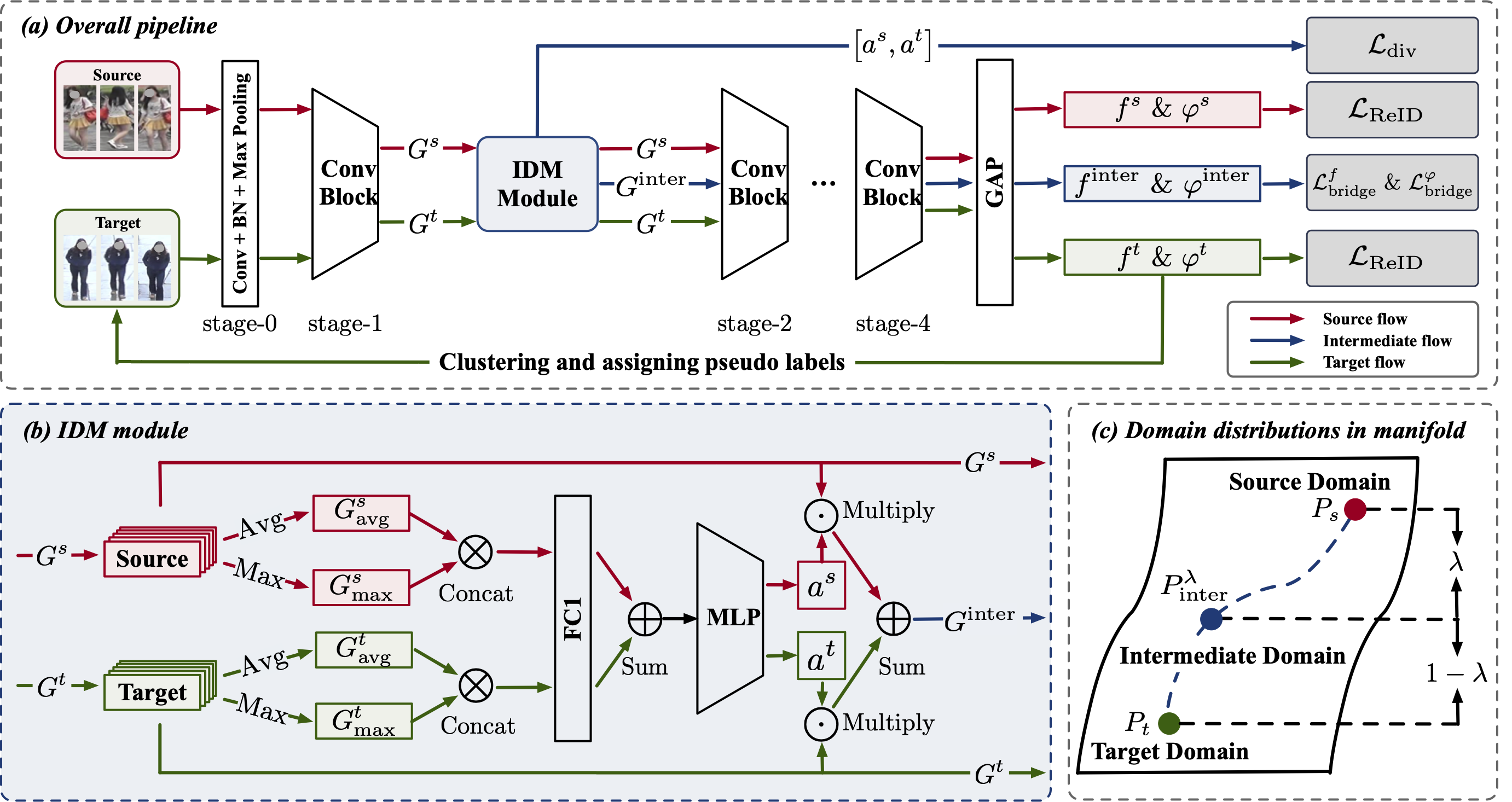}
\end{center}
\caption{Illustration of our method. 
(a) In a joint training way, our proposed IDM module can be plugged after any stage (\textit{e.g.,} stage-1) of ResNet-50, where the source and target domains' hidden representations (\textit{i.e.,} $G^{s}, G^{t}$) are mixed with IDM to generate intermediate domains' representations $G^{\rm inter}$. The $G^{s}$, $G^{t}$, and $G^{\rm inter}$ are together fed into the next stage of the same network and the final outputs are their features and predictions: $(f^{s},\varphi^{s})$, $(f^{t},\varphi^{t})$, and $(f^{\rm inter},\varphi^{\rm inter})$.
GAP means the global average pooling layer.
For simplicity, we omit the classifier. The common ReID loss $\mathcal L_{\rm ReID}$ (including the classification loss $\mathcal L_{\rm cls}$ and triplet loss $\mathcal L_{\rm tri}$) are enforced on the source and target domains. To generate more appropriate intermediate domains, we enforce bridge losses ($\mathcal L^{f}_{\rm bridge}$ and $\mathcal L^{\varphi}_{\rm bridge}$) on intermediate domains' $f^{\rm inter}$ and $\varphi^{\rm inter}$ respectively, and enforce a diversity loss ($\mathcal L_{\rm div}$) to regularize the domain factors $[a^{s}, a^{t}]$ obtained from the IDM module. (b) The detailed structure of the IDM module. ``Avg'' and ``Max'' mean average-pooling and max-pooling operations respectively. (c)~Assuming different domains as different points in a manifold. An appropriate intermediate domain $P^{\lambda}_{\rm inter}$ should be located along with the shortest geodesic path between the source and target domains (\textit{i.e.,} $P_{s}$ and $P_{t}$), making $P^{\lambda}_{\rm inter}$ keep the right distance to $P_{s}$ and $P_{t}$.
}
\label{fig:pipeline}
\vspace{-1em}
\end{figure*}

\textbf{Unsupervised Domain Adaptative Person Re-ID.} In recent years, many UDA re-ID methods have been proposed and they can be mainly categorized into three types based on their training schemes, \textit{i.e.,} GAN transferring \cite{wei2018person,deng2018image,huang2019sbsgan,zou2020joint}, fine-tuning \cite{song2020unsupervised,fu2019self,ge2020mutual,dai2020dual,chen2020deep,zhai2020multiple,jin2020global,lin2020unsupervised,zhai2020ad}, and joint training \cite{zhong2019invariance,zhong2020learning,wang2020unsupervised,ge2020self,ding2020adaptive}. 
GAN transferring methods use GANs to transfer images' style across domains \cite{wei2018person,deng2018image} or disentangle features into id-related/unrelated features \cite{zou2020joint}.
For fine-tuning methods, they first train the model with labeled source data and then fine-tune the pre-trained model on target data with pseudo labels. 
The key component of these methods is how to alleviate the effects of the noisy pseudo labels. 
However, these methods ignore the labeled source data while fine-tuning on the target data, which will hinder the domain adaptation process because of the catastrophic forgetting in networks. For joint training methods, they combine the source and target data together and train on an ImageNet-pretrained network from scratch. 
All these joint training methods often utilize the memory bank \cite{xiao2017joint,wu2018unsupervised} to improve target domain features' discriminability. However, these methods just take both the source and target data as the network's input and train jointly while neglecting the bridging between both domains, \textit{i.e.,} what information of the two domains' dissimilarities/similarities can be utilized to improve features' discriminability in UDA re-ID.
Different from all the above UDA re-ID methods, we propose to consider the bridging between the source and target domains by modeling appropriate intermediate domains with a plug-and-paly module, which is helpful for gradually adapting between two extreme domains in UDA re-ID.

\textbf{Mixup and Variants.} Mixup \cite{zhang2017mixup} is an effective regularization technique to improve the generalization of deep networks by linearly interpolating the image and label pairs, where the interpolating weights are randomly sampled from a Dirichlet distribution. Manifold Mixup \cite{verma2019manifold} extends Mixup to a more general form which can linearly interpolate data at the feature level. Recently, Mixup has been applied to many tasks like point cloud classification \cite{chen2020pointmixup}, object detection \cite{zhang2019bag}, and closed-set domain adaptation\cite{xu2020adversarial,wu2020dual,na2021fixbi}. Our work differs from these Mixup variants in: (1) All the above methods take Mixup as a data/feature augmentation technique to improve models' generalization, while we bridge two extreme domains by generating intermediate domains for UDA re-ID. (2) We design an IDM module and enforce specific losses on it to control the bridging process while all the above methods often linearly interpolate data using the random interpolation ratio without constraints.

\section{Proposed Method}

The purpose of our method is to generate appropriate intermediate domains to bridge the source and target domains. By training the source, target, and intermediate domains together at the same time, the network will adaptively characterize the distributions of the source and target domains. This will smoothly adapt between the two extreme domains and better transfer the source knowledge to improve the model's performance on the target domain.
\subsection{Overview}
In UDA re-ID, we are often given a labeled source domain dataset $\{(x^{s}_{i},y^{s}_{i})\}$ and an unlabeled target domain dataset $\{x^{t}_{i}\}$. The source dataset contains $N^{s}$ labeled person images, and the target dataset contains $N^{t}$ unlabeled images. Each source image $x_{i}^{s}$ is associated with a person identity $y_{i}^{s}$ and the total number of source domain identities is $C^{s}$. We use ResNet-50 \cite{he2016deep} as the backbone network $f(\cdot)$ and add a hybrid classifier $\varphi(\cdot)$ after the global average pooling (GAP) layer, where the hybrid classifier is comprised of the batch normalization layer and a $C^{s}+C^{t}$ dimensional fully connected (FC) layer followed by a softmax activation function. $C^{t}$ is the cluster number when performing DBSCAN clustering on the target domain features at the beginning of every training epoch like existing clustering-based UDA re-ID methods \cite{fu2019self,song2020unsupervised,ge2020self}. After clustering, we assign the pseudo labels $y^{t}_{i}$ for the target data $x^{t}_{i}$. 

Figure \ref{fig:pipeline} \textcolor{red}{(a)} shows the pipeline of our method. The overall training scheme is jointly training on the source and target data. A mini-batch including $n$ source samples and $n$ target samples is fed to the network. Our proposed Intermediate Domain Module (IDM) can be plugged after any stage of the deep network like ResNet-50. As shown in Figure \ref{fig:pipeline}~\textcolor{red}{(a)}, the IDM module is plugged between the stage-1 and stage-2, where the source and target domains' hidden representations are mixed with the IDM module to generate new intermediate domains' representations. Next, all the representations of the source, target, and intermediate domains are fed into the stage-2 until the end of the network. We enforce ReID loss $\mathcal L_{\rm ReID}$ (including the classification loss \cite{zheng2016person} and triplet loss \cite{hermans2017defense}) on the source and target domains' features ($f^{s}, f^{t}$) and predictions ($\varphi^{s},\varphi^{t}$). To make the IDM module generate more appropriate intermediate domain representations, we propose two bridge losses and a diversity loss. The bridge losses ($\mathcal L_{\rm bridge}^{f}, \mathcal L_{\rm bridge}^{\varphi}$)  are respectively enforced on intermediate domains' features $f^{\rm inter}$ and predictions $\varphi^{\rm inter}$. The diversity loss $\mathcal L_{\rm div}$ is used to enforce on the domain factors obtained from the IDM module, aiming to prevent the module from being over-fitting to either of two extreme domains. The IDM module is only used for training and will be discarded in testing.

\subsection{Intermediate Domain Module}

Our proposed IDM module can be plugged after the hidden stage in the backbone network. 
The module takes both source and target hidden stage's representations as the input and generate two domain factors \textit{i.e.,} $a^{s}$ and $a^t$. With these two domain factors, we can mix the source and target domains' hidden representations to generate intermediate domains' representations on-the-fly.

Specifically, we denote backbone network as $f(x)=f_{m}(g_{m}(x))$, where $g_{m}$ represents the part of the network mapping the input data $x$ to the hidden representation $g_{m}(x)\in \mathbb{R}^{h\times w\times c}$ (\textit{i.e.,} feature map)
after the $m$-th stage and $f_{m}$ represents the part of the network mapping the hidden representation $g_{m}(x)$ to the 2048-dim feature after the GAP layer. As shown in Figure~\ref{fig:pipeline} \textcolor{red}{(b)}, the IDM module is composed of a fully connected layer FC1, and a MLP (Muti-Layer Perception) followed by a softmax function.

In a mini-batch including $n$ source and $n$ target samples, we randomly combine two domains' samples into $n$ pairs.
For each sample pair $(x^{s},x^{t})$, we can obtain their feature maps at the $m$-th hidden stage: $G^{s}, G^{t} \in \mathbb{R}^{h\times w\times c}$, both of which are performed with the average-pooling and max-pooling operations respectively, generating the $1\times 1 \times c$ dimensional features \textit{i.e.,} ($G_{avg}^{s}, G_{max}^{s}$) for the source domain, and ($G_{avg}^{t}, G_{max}^{t}$) for the target domain. We concatenate the $avg$ and $max$ features for each domain and feed them to FC1. The output feature vectors of FC1 are merged using element-wise summation and fed into MLP to obtain a domain factor vector $a\in \mathbb{R}^{2}$. The procedure to obtain the domain factors in Figure \ref{fig:pipeline} \textcolor{red}{(b)} is as follows:
\begin{equation}
\small
a=\delta(MLP(FC1([G_{avg}^{s};G_{max}^{s}])+FC1([G_{avg}^{t};G_{max}^{t}]))),
\label{eq: attention}
\end{equation}
where $\delta(\cdot)$ is the softmax function.
Next, we define the operation of generating intermediate domain representations after the $m$-th hidden stage by mixing the source and target representations with two domain factors as follows:
%\vspace{-0.5em}
\begin{equation}
\small
    G^{\rm inter}=a^{s}\cdot G^{s}+a^{t}\cdot G^{t},
\label{eq:Mix}
\end{equation}
where $G^{s}$ and $G^{t}$ are the $m$-th hidden stage's representations, and $a^{s}$ and $a^{t}$ are the two domain factors for the source and target domains respectively ($[a^{s},a^{t}]=a$).

\subsection{Modelling Appropriate Intermediate Domains}

We can generate infinite intermediate domains by the IDM module in a mixing operation as Eq. (\ref{eq:Mix}). However, only a part of these intermediate domains can be appropriate to bridge the source and target domains \cite{gopalan2013unsupervised}. Inspired by the traditional works \cite{gong2012geodesic,gopalan2013unsupervised} that construct intermediate domains using kernel-based methods, all domains can be assumed to be located in a manifold. Only those intermediate domains located along the shortest geodesic path between the source and target domains can be helpful for the smooth adaptation procedure \cite{gopalan2013unsupervised}. Based on this ``shortest geodesic path'' definition, we argue that appropriate intermediate domains should satisfy two properties and further propose specifically-designed losses to ensure the two properties in the concept of UDA re-ID. We
denote distributions of source, target, and intermediate domains as $P_{s}$, $P_{t}$, and $P_{\rm inter}$, and use $d(\cdot)$ to represent geodesic distance in a manifold. As shown in Figure \ref{fig:pipeline} \textcolor{red}{(c)}, different domains can be seen as different points in a manifold.

\textbf{Definition (Shortest geodesic path):} \textit{Appropriate intermediate domains should be located onto the shortest geodesic path between the source and target domains, i.e., $d(P_{s},P_{\rm inter})+d(P_{t},P_{\rm inter})=d(P_{s},P_{t})$.}

Intuitively, for an intermediate domain ${P}'_{\rm inter}$ not on the shortest geodesic path, the triangle inequality ensures that $d(P_{s},{P}'_{\rm inter})+d(P_{t},{P}'_{\rm inter})>d(P_{s},P_{t})$. As in UDA re-ID, our goal is to reduce the distribution shift $d(P_{s},P_{t})$ between the source and target domains and the inappropriate intermediate domain ${P}'_{\rm inter}$ can bring about ``extra domain shifts'' (\textit{i.e.,} the above inequality) that will degenerate the adaptation procedure.

\textbf{Property 1 (Distance should be proportional):}
\textit{We use a variable $\lambda\in [0,1]$ to control the distance between the source domain to any intermediate domain $P^{(\lambda)}_{\rm inter}$ on the shortest geodesic path, i.e., $d(P_{s},P^{(\lambda)}_{\rm inter})=\lambda\cdot d(P_{s},P_{t})$. 
The distances satisfy the proportional relationship:}
\begin{equation}
\small
\frac{d(P_{s},P^{(\lambda)}_{\rm inter})}{d(P_{t},P^{(\lambda)}_{\rm inter})}=\frac{\lambda}{1-\lambda}.
\label{eq:proportional}
\end{equation}

\textbf{Sketch of Proof:} From the shortest geodesic path definition, we can derive $d(P_{s},P^{(\lambda)}_{\rm inter})+d(P_{t},P^{(\lambda)}_{\rm inter})=d(P_{s},P_{t})$. Then from $d(P_{s},P^{(\lambda)}_{\rm inter})=\lambda\cdot d(P_{s},P_{t})$, we can get the proportional relationship as Eq. (\ref{eq:proportional}).

As shown in Eq. (\ref{eq:proportional}), if $\lambda=0$ (or $\lambda$=1), the intermediate domain $P^{(\lambda)}_{\rm inter}$ is identical to the source (or target) domain $P_{s}$ (or $P_{t}$). 
In our method, we use the domain factors $a$ to model the proportional relationship.
As shown in Eq. (\ref{eq:Mix}), the generated intermediate domains' representations depend on two domain factors: $a^{s}$ and $a^{t}$, where $a^{s},a^{t}\in [0,1]$ and $a^{s}+a^{t}=1$ as we use the softmax function $\delta$ in Eq.~(\ref{eq: attention}). The domain factors $a$ can be seen as the relevance of the intermediate domain to the other two extreme domains. 
%For example, if $a^{s}=1$ (or $a^{t}=1$), the intermediate domain is identical to the source (or target). 
Thus, the distance relationship (contrary to the relevance relationship) between intermediate domains and other two domains can be formulated as follows:
\begin{equation}
\small
%\vspace{-0.3em}
\frac{d(P_{s},P^{(a)}_{\rm inter})}{d(P_{t},P^{(a)}_{\rm inter})}=\frac{a^{t}}{a^{s}}=\frac{1-a^{s}}{a^{s}}=\frac{a^{t}}{1-a^{t}},
\label{eq:proportional2}
%\vspace{-0.5em}
\end{equation}
where $P^{(a)}_{\rm inter}$ is the intermediate domain controlled by the domain factors $a$.  When $a^{s}$ gets close to 1, the relevance between $P^{(a)}_{\rm inter}$ and $P_{s}$ becomes larger while the distance $d(P_{s},P^{(a)}_{\rm inter})$ becomes smaller. 

Based on the above analysis, generating appropriate intermediate domains with domain factors $a$ becomes finding the points located along the shortest path that should be closest to both $P_{s}$ and $P_{t}$, and satisfy Eq. (\ref{eq:proportional2}). Thus, the problem can be converted into minimizing the loss as:
\begin{equation}
\small
\mathcal L_{\rm bridge}=a^{s}\cdot d(P_{s},P^{(a)}_{\rm inter})+a^{t}\cdot d(P_{t},P^{(a)}_{\rm inter}).
\label{eq:loss_mix}
%\vspace{-0.5em}
\end{equation}
If $a^{s}>a^{t}$, $\mathcal L_{\rm bridge}$ will penalize more on $d(P_{s},P^{(a)}_{\rm inter})$, otherwise it will penalize more on $d(P_{t},P^{(a)}_{\rm inter})$. Intermediate domains can utilize domain factors $a$ to adaptively balance the minimization of distribution shifts between the source and target domains, resulting in a smooth domain adapattion procedure.

In a deep model for UDA re-ID, we consider to enforce the bridge loss (Eq. (\ref{eq:loss_mix})) on both the prediction and feature spaces of intermediate domains. For the prediction space, we use the cross-entropy to measure the distribution gap between intermediate domains' prediction logits and other two extreme domains' (pseudo) labels (Eq. (\ref{eq:loss_mix_label})). For the feature space, we use the L2-norm to measure features' distance among domains (Eq. (\ref{eq:loss_mix_feat})). Our proposed two bridge losses are formulated as follows:
\begin{equation}
\small
\mathcal L_{\rm bridge}^{\varphi}=-\frac{1}{n}\sum_{i=1}^{n}\sum_{k\in \{s,t\}}a_{i}^{k}\cdot \left [y_{i}^{k}log(\varphi(f_{m}(G_{i}^{\rm inter})))\right ],
\label{eq:loss_mix_label}
%\vspace{-1.5em}
\end{equation}
\begin{equation}
\small
\mathcal L_{\rm bridge}^{f}=\frac{1}{n}\sum_{i=1}^{n}\sum_{k\in \{s,t\}}a_{i}^{k}\cdot \left \|f_{m}(G^{k}_{i}) -f_{m}(G_{i}^{\rm inter}) \right \|_{2}.
\label{eq:loss_mix_feat}
\end{equation}
In Eq.~(\ref{eq:loss_mix_label}) (\ref{eq:loss_mix_feat}), we use $k$ to indicate the domain (source or target) and use $i$ to index the data in a mini-batch. $G^{k}_{i}$ is the $k$ domain's representation at the $m$-th stage.
$G_{i}^{\rm inter}$ is the intermediate domain's representation at the $m$-th hidden stage by mixing $G_{i}^{s}$ and $G_{i}^{t}$ as in Eq. (\ref{eq:Mix}).
The $f_{m}(\cdot)$ is the mapping from the $m$-th stage to the features after GAP layer and $\varphi(\cdot)$ is the classifier.

\textbf{Property2 (Diversity):}
\textit{Intermediate domains should be as diverse as possible.}

\textbf{Sketch of Proof:} Given any intermediate domain $P^{(\lambda)}_{\rm inter}$ along the shortest geodesic path from \textbf{Property 1}, the intermediate domain will be mainly dominated by the source (or target) domain if $\lambda\in [0,0.5)$ (or $\lambda\in (0.5,1]$). Neither conditions can well bridge the source and target domains, \textit{i.e.,} the shortest geodesic path will be cut off.

To ensure the above property, we propose the diversity loss to enforce on the domain factors $a$ by maximizing their differences within a mini-batch. This loss is as follows:
\begin{equation}
\small
\mathcal L_{\rm div}=-[\sigma(\{a^{s}_{i}\}_{i=1}^{n})+\sigma(\{a^{t}_{i}\}_{i=1}^{n})],
%\vspace{-1em}
\label{eq:loss_div}
\end{equation}
where $\sigma(\cdot)$ means calculating the standard deviation of the values in a mini-batch. By minimizing $\mathcal L_{\rm div}$, we can enforce intermediate domains to be as diverse as enough to model the characteristics of the ``shortest geodesic path'', which can better bridge the source and target domains.

In conclusion, we use $\mathcal L_{\rm bridge}^{\varphi}$ and $\mathcal L_{\rm bridge}^{f}$ to enforce generated intermediate domains to be located along the ``shortest geodesic path'' and use $\mathcal L_{\rm div}$ to characterize more intermediate domains, which will effectively build a bridge between the two extreme domains. This bridge can adaptively balance the adaptation between the source and target domains when training our network over both the source and target data, \textit{i.e.,} the model can gradually minimize the discrepancy between the two extreme distributions.

\subsection{Overall Training}
The ResNet-50 backbone contains five stages, where stage-0 is comprised of the first Conv, BN and Max Pooling layer and stage-1/2/3/4 correspond to the other four convolutional blocks.
We plug our proposed IDM module after the $m$-th stage of ResNet-50 (\textit{i.e.,} stage-0/1/2/3/4 as shown in Figure \ref{fig:pipeline}). The overall training loss is as follows:
%\vspace{-0.5em}
\begin{equation}
\small
    \mathcal L= \mathcal L_{\rm ReID}+\mu_{1}\cdot \mathcal L_{\rm bridge}^{\varphi}+\mu_{2}\cdot \mathcal L_{\rm bridge}^{f}+\mu_{3}\cdot L_{\rm div},
\label{eq:overall_loss}
\end{equation}
where $\mathcal L_{\rm ReID} = (1-\mu_{1}) \cdot \mathcal L_{\rm cls} + \mathcal L_{\rm tri}$, and $\mu_{1}, \mu_{2}, \mu_{3}$ are the weights to balance losses. The training and testing procedure is shown in Algorithm \ref{alg:algorithm1} in the supplementary material.

\section{Experiments}
Five datasets including Market-1501 \cite{zheng2015scalable}, DukeMTMC-reID \cite{ristani2016performance,zheng2017unlabeled}, MSMT17 \cite{wei2018person}, PersonX \cite{sun2019dissecting}, and Unreal \cite{zhang2021unrealperson} are used in our experiments. 
We use mean average precision (mAP) and Rank-1/5/10 (R1/5/10) of CMC to evaluate performances. In training, we do not use any additional information like temporal consistency in JVTC+ \cite{li2020joint}.
In testing, there are no post-processing techniques like re-ranking \cite{zhong2017re} or multi-query fusion \cite{zheng2015scalable}. Implementation details can be seen in the supplementary material.

\subsection{Ablation Study}

In this section, we will evaluate the effectiveness of different components of our method. The ``Oracle'' method in Table~\ref{tab:ablation} means we use the target domain's ground-truth to train together with the source domain, which can be seen as the upper bound of the performance for UDA re-ID in a joint training pipeline. 

\textbf{Effectiveness of our proposed IDM module on different baseline methods.} We propose two baseline methods including Naive Baseline (Baseline1) and Strong Baseline (Baseline2). Naive Baseline means only using $\mathcal L_{\rm ReID}$ to train the source and target data in joint manner as shown in Figure \ref{fig:pipeline}.
Compared with Naive Baseline, Strong Baseline uses XBM \cite{wang2020cross} to mine more hard negatives for the triplet loss, which is a variant of the memory bank \cite{xiao2017joint,wu2018unsupervised} and easy to implement. 
Similar to those UDA re-ID methods \cite{zhong2020learning,ge2020self} using the memory bank, we set the memory bank size as the number of all the training data.
As shown in Table \ref{tab:ablation}, no matter which baseline we use, our methods can obviously outperform the baseline methods by a large margin. Take Market$\to$Duke as an example, mAP/R1 of Baseline2+our IDM (full) is 4.7\%/3.5\% higher than Baseline2, and mAP/R1 of Baseline1+our IDM (full) is 5.2\%/3.9\% higher than Baseline1. Because of many state-of-the-arts methods \cite{ge2020self,zheng2020exploiting,li2020joint,zheng2021group} use the memory bank to improve the performance on the target domain, we use Strong Baseline to implement our IDM for fairly comparing with them.

\begin{table}[tp]
\begin{center}
\caption{Ablation studies on different components of our method. Baseline1 (Naive Baseline): only using $\mathcal L_{\rm ReID}$ to train the source and target domains jointly. Baseline2 (Strong Baseline): Baseline1 + XBM \cite{wang2020cross}. D/M: Duke/Market.}
\label{tab:ablation}
\footnotesize
\begin{tabular}{l|cc|cc}
\hline
\multicolumn{1}{c|}{\multirow{2}{*}{Method}} & \multicolumn{2}{c|}{D $\to$ M} & \multicolumn{2}{c}{M $\to$ D} \\ \cline{2-5} 
\multicolumn{1}{c|}{} & mAP & R1 & mAP & R1 \\ \hline
Oracle &83.9 &93.2  &75.0 &86.1  \\ 
Baseline1  &77.0  &90.6 &63.4 &78.4  \\ %\hline
Baseline1 + Our IDM w/o $\mathcal L_{\rm bridge}^{\varphi}$ &79.4 &91.5 &66.2 &79.8  \\
Baseline1 + Our IDM w/o $\mathcal L_{\rm bridge}^{f}$ &79.2 &91.2 &65.8 &80.4  \\
Baseline1 + Our IDM w/o $\mathcal L_{\rm div}$ &78.9 &91.1 &64.8 &79.3  \\ %\hline
Baseline1 + Our IDM (full) &81.9 &92.4 &68.6 &82.3  \\ \hline \hline
Oracle + XBM &86.9 &94.8 &78.1 &88.3 \\
Baseline2  &79.1  &91.2  &65.8 &80.1  \\ %\hline
Baseline2 + Our IDM w/o $\mathcal L_{\rm bridge}^{\varphi}$ &80.7 &92.0 &67.3 &81.8  \\
Baseline2 + Our IDM w/o $\mathcal L_{\rm bridge}^{f}$ &81.6 &92.2 &69.0 &82.0 \\
Baseline2 + Our IDM w/o $\mathcal L_{\rm div}$ &79.7 &91.6 &67.8 &81.6  \\ %\hline
Baseline2 + Our IDM (full) &82.8 &93.2 &70.5 &83.6  \\ \hline
\end{tabular}
\end{center}
\vspace{-1em}
\end{table}

\textbf{Effectiveness of the bridge losses for the prediction and feature spaces.} The bridge losses are enforced on both the prediction and feature spaces of intermediate domains. With both losses, we can enforce the generated intermediate domains to be located onto the geodesic path between the source and target domains in a manifold, which are more appropriate to bridge domains to perform more smooth domain adaptation. Take Market$\to$Duke as an example, mAP/R1 of Baseline2+IDM w/o $\mathcal L_{\rm bridge}^{\phi}$ is 3.2\%/1.8\% lower than Baseline2+IDM, and mAP/R1 of Baseline2+IDM w/o $\mathcal L_{\rm bridge}^{f}$ is 1.5\%/1.6\% lower than Baseline2+IDM. From Table \ref{tab:ablation}, the method w/o $\mathcal L_{\rm bridge}^{f}$ is superior to that w/o $\mathcal L_{\rm bridge}^{\phi}$ for Baseline2, while the conclusion is just opposite for Baseline1. The reason may be that Baseline2 uses XBM to mine harder negatives in the feature space, which will affect the effectiveness of~$\mathcal L_{\rm bridge}^{f}$.

\textbf{Effectiveness of the diversity loss.} We use the diversity loss $\mathcal L_{\rm div}$ to prevent the IDM module from being over-fitting to either of the source and target domains when generating intermediate domains. Baseline2+IDM w/o $\mathcal L_{\rm div}$ degenerates much when comparing with the full method. Take Market $\to$ Duke as an example (Table \ref{tab:ablation}), mAP of Strong Baseline+IDM w/o  $\mathcal L_{\rm div}$ is 2.7\% lower than the full method.

\begin{table}[tp]
\begin{center}
\caption{Study about which stage of ResNet-50 to plug the IDM module.}
\label{tab:stage}
\footnotesize
\begin{tabular}{l|cc|cc}
\hline
\multicolumn{1}{c|}{\multirow{2}{*}{Method}} & \multicolumn{2}{c|}{Duke $\to$ Market} & \multicolumn{2}{c}{Market $\to$ Duke} \\ \cline{2-5} 
\multicolumn{1}{c|}{} & mAP & R1 & mAP & R1 \\ \hline
Strong Baseline &79.1  &91.2  &65.8  &80.1  \\ \hline
IDM after stage-0 &\textbf{82.8}  &\textbf{93.2} &\textbf{70.5} &\textbf{83.6} \\
IDM after stage-1 & 82.3 & 92.2 &70.3 &83.2 \\
IDM after stage-2 & 81.9 & 92.0 &69.2	&82.3 \\
IDM after stage-3 & 81.3 & 91.8 &68.6 &82.1  \\
IDM after stage-4 & 80.2 & 91.3 &68.4 &81.5  \\ \hline
\end{tabular}
\end{center}
\vspace{-1.5em}
\end{table}

\textbf{Effectiveness on which stage to plug the IDM module.} Our IDM module is a plug-and-play module which can be plugged after any stage of the backbone network. In our experiments, we use ResNet-50 as the backbone, which has five stages: stage-0 is comprised of the first Conv, BN and Max Pooling layer and stage-1/2/3/4 correspond to the other four convolutional blocks. We plug our IDM module after different stages to study how different stages will affect the performance of the IDM module. No matter after which stage, our method can all outperform the strong baseline.
As shown in Table \ref{tab:stage}, the performance of the IDM module is gradually declining with the deepening of the network. This phenomenon shows that the domain gap becomes larger and the transferable ability becomes weaker at higher/deeper layers of the network, which satisfies the theory of the domain adaptation~\cite{long2018transferable}. Based on the above analyses, we plug the IDM module after stage-0 of the backbone in all our experiments if not specified.

\begin{figure}[tp]
    \centering
    \includegraphics[width=1\linewidth]{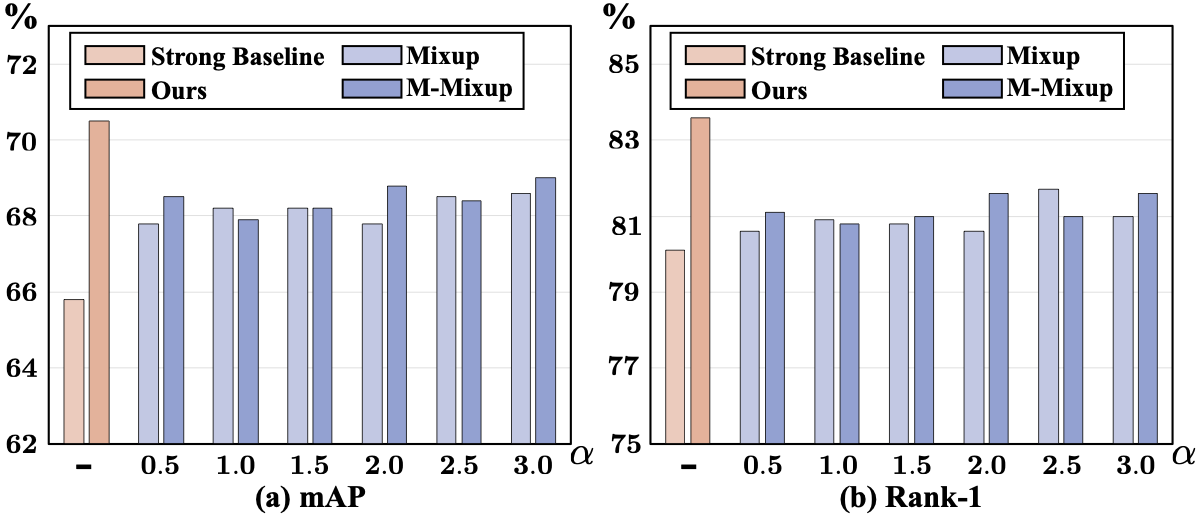}
    \caption{Comparison with traditional mixup methods (Mixup \cite{zhang2017mixup}, M-Mixup \cite{verma2019manifold}) on Market$\to$Duke. The interpolation ratio in these mixup methods is randomly sampled from a beta distribution ${\rm Beta}(\alpha,\alpha)$.}
    \label{fig:vs_mixup}
    \vspace{-2em}
\end{figure}

\textbf{Ours vs. traditional mixup methods.}
We compare our method with the traditional Mixup \cite{zhang2017mixup} and Manifold Mixup (M-Mixup) \cite{verma2019manifold} methods in Figure \ref{fig:vs_mixup}. We use Mixup to randomly mix the source and target domains at the image-level, and use M-Mixup to randomly mix the two domains at the feature-level. The interpolation ratio in Mixup and M-Mixup is randomly sampled from a beta distribution ${\rm Beta}(\alpha,\alpha)$. Specifically, Mixup and M-Mixup are only the image/feature augmentation technology and we use them to randomly mix the source and target domains. Different from them, our method can adaptively bridge the source and target domains to ensure the two properties that appropriate intermediate domains should satisfy. Thus, our method is superior to these mixup methods in UDA re-ID.

\textbf{Scalability for different backbones.} Our proposed IDM module can be easily scalable to other backbones, such as IBN-ResNet-50 \cite{pan2018two}. As shown in Table \ref{tab:resnet-ibn}, the performance gain is especially obvious on the largest and challenging MSMT dataset. In addition to IBN \cite{pan2018two}, our method can also be easily implemented with other normalization techniques like AdaBN~ \cite{li2016revisiting} and CBN \cite{zhuang2020rethinking}.

\begin{table}[tp]
\begin{center}
\caption{Comparison with different backbones in our method.
}
\label{tab:resnet-ibn}
\footnotesize
\begin{tabular}{l|l|cc|cc}
\hline
\multicolumn{1}{c|}{\multirow{2}{*}{Source}} & \multicolumn{1}{c|}{\multirow{2}{*}{Target}} & \multicolumn{2}{c|}{w/ ResNet-50} & \multicolumn{2}{c}{w/ IBN-ResNet-50} \\ \cline{3-6} 
\multicolumn{1}{c|}{} & \multicolumn{1}{c|}{} & mAP & R1 & mAP & R1 \\ \hline
Duke &Market  &82.8  &93.2  &\textbf{83.9} &\textbf{93.4}  \\
Market &Duke  &70.5 &83.6  &\textbf{71.1} &\textbf{83.9}  \\
Duke &MSMT  &35.4  &63.6  &\textbf{40.8}  &\textbf{69.6}  \\
Market &MSMT  &33.5  &61.3  &\textbf{39.3}  &\textbf{67.2}  \\ \hline
\end{tabular}
\end{center}
\vspace{-2.5em}
\end{table}

\subsection{Comparison with the State-of-the-arts}

The existing state-of-the-arts (SOTA) UDA re-ID works commonly evaluate the performance on four real $\to$ real tasks \cite{fu2019self,zhong2020learning}. Recently, more challenging synthetic $\to$ real tasks \cite{ge2020self,zhang2021unrealperson} are proposed, where they use the synthetic dataset PersonX \cite{sun2019dissecting} or Unreal \cite{zhang2021unrealperson} as the source domain and test on other three real re-ID datasets. 
Using ResNet-50 as the backbone, all the results in Table \ref{tab:SOTA-real2real},~\ref{tab:SOTA-synthetic2real} show that our method can outperform the SOTAs significantly.
\begin{table*}[htp]
\caption{Comparison with the state-of-the-art UDA re-ID methods on real $\to$ real tasks.
}
\footnotesize
\label{tab:SOTA-real2real}
\begin{center}
\begin{tabular}{l|c|p{1cm}<{\centering}p{1cm}<{\centering}p{1cm}<{\centering}p{1cm}<{\centering}|p{1cm}<{\centering}p{1cm}<{\centering}p{1cm}<{\centering}p{1cm}<{\centering}}
\hline
\multirow{2}{*}{Methods} & \multirow{2}{*}{Reference} & \multicolumn{4}{c|}{DukeMTMC-ReID $\to$ Market-1501} & \multicolumn{4}{c}{Market-1501 $\rightarrow$ DukeMTMC-ReID} \\ \cline{3-10} 
                         &                            & mAP      & R1       & R5       & R10     & mAP      & R1       & R5      & R10     \\ \hline 
PUL \cite{fan2018unsupervised}                     & TOMM 2018                  & 20.5     & 45.5     & 60.7     & 66.7    & 16.4     & 30.0     & 43.4    & 48.5    \\
TJ-AIDL \cite{wang2018transferable}                 & CVPR 2018                  & 26.5     & 58.2     & 74.8     & 81.1    & 23.0     & 44.3     & 59.6    & 65.0    \\
SPGAN+LMP \cite{deng2018image}                & CVPR 2018                  & 26.7     & 57.7     & 75.8     & 82.4    & 26.2     & 46.4     & 62.3    & 68.0    \\
HHL \cite{zhong2018generalizing}                     & ECCV 2018                  & 31.4     & 62.2     & 78.8     & 84.0    & 27.2     & 46.9     & 61.0    & 66.7    \\
ECN \cite{zhong2019invariance}                     & CVPR 2019                  & 43.0     & 75.1     & 87.6     & 91.6    & 40.4     & 63.3     & 75.8    & 80.4    \\
PDA-Net \cite{li2019cross}                 & ICCV 2019                  & 47.6     & 75.2     & 86.3     & 90.2    & 45.1     & 63.2     & 77.0    & 82.5    \\
PCB-PAST \cite{zhang2019self}                & ICCV 2019                  & 54.6     & 78.4     & -        & -       & 54.3     & 72.4     & -       & -       \\
SSG \cite{fu2019self}                     & ICCV 2019                  & 58.3     & 80.0     & 90.0     & 92.4    & 53.4     & 73.0     & 80.6    & 83.2    \\
MMCL \cite{wang2020unsupervised}                 & CVPR 2020                 & 60.4     & 84.4     & 92.8     & 95.0    & 51.4     & 72.4     & 82.9    & 85.0    \\
ECN-GPP \cite{zhong2020learning}                 & TPAMI 2020                 & 63.8     & 84.1     & 92.8     & 95.4    & 54.4     & 74.0     & 83.7    & 87.4    \\
JVTC+ \cite{li2020joint} &ECCV 2020 &67.2 &86.8 &95.2 &97.1 &66.5 &80.4 &89.9 &92.2  \\
AD-Cluster \cite{zhai2020ad}              & CVPR 2020                  & 68.3     & 86.7     & 94.4     & 96.5    & 54.1     & 72.6     & 82.5    & 85.5    \\
MMT \cite{ge2020mutual}                     & ICLR 2020                  & 71.2     &87.7     &94.9     &96.9    &65.1     &78.0     &88.8    &92.5    \\ 
CAIL \cite{luo2020generalizing} &ECCV 2020 &71.5 &88.1 &94.4 &96.2 &65.2 &79.5 &88.3 &91.4 \\
NRMT \cite{zhao2020unsupervised} &ECCV 2020  &71.7 &87.8 &94.6 &96.5 &62.2 &77.8 &86.9 &89.5 \\
MEB-Net \cite{zhai2020multiple} &ECCV 2020 &76.0 &89.9 &96.0 &97.5 &66.1 &79.6 &88.3 &92.2 \\ 
SpCL \cite{ge2020self} &NeurIPS 2020 &76.7 &90.3 &96.2 &97.7 &68.8 &\underline{82.9} &90.1 &92.5 \\ 
Dual-Refinement \cite{dai2020dual} &TIP 2021 &78.0 &90.9 &96.4 &97.7 &67.7 &82.1 &90.1 &92.5 \\
UNRN \cite{zheng2020exploiting} &AAAI 2021 &78.1 &91.9 &96.1 &\underline{97.8} &69.1 &82.0 &\underline{90.7} &\underline{93.5}    \\ 
GLT \cite{zheng2021group} &CVPR 2021 &\underline{79.5} &\underline{92.2} &\underline{96.5} &\underline{97.8} &\underline{69.2} &82.0 &90.2 &92.8    \\ \hline
IDM (Ours) & ICCV 2021  &\textbf{82.8} &\textbf{93.2} &\textbf{97.5} &\textbf{98.1} &\textbf{70.5} &\textbf{83.6} &\textbf{91.5} &\textbf{93.7}  \\ \hline \hline

\multirow{2}{*}{Methods} & \multirow{2}{*}{Reference} & \multicolumn{4}{c|}{Market-1501 $\to$ MSMT17} & \multicolumn{4}{c}{DukeMTMC-reID $\to$ MSMT17} \\ \cline{3-10} 
 &  & mAP & R1 & R5 & R10 & mAP & R1 & R5 & R10 \\ \hline
ECN \cite{zhong2019invariance} & CVPR 2019 & 8.5 & 25.3 & 36.3 & 42.1 & 10.2 & 30.2 & 41.5 & 46.8 \\
SSG \cite{fu2019self} & ICCV 2019 & 13.2 & 31.6 & - & 49.6 & 13.3 & 32.2 & - & 51.2 \\
ECN-GPP \cite{zhong2020learning} & TPAMI 2020 & 15.2 & 40.4 & 53.1 & 58.7 & 16.0 & 42.5 & 55.9 & 61.5 \\
MMCL \cite{wang2020unsupervised} & CVPR 2020 & 15.1 & 40.8 & 51.8 & 56.7 & 16.2 & 43.6 & 54.3 & 58.9 \\
NRMT \cite{zhao2020unsupervised} &ECCV 2020 &19.8 &43.7 &56.5 &62.2 &20.6 &45.2 &57.8 &63.3 \\
CAIL \cite{luo2020generalizing} &ECCV 2020 &20.4 &43.7 &56.1 &61.9 &24.3 &51.7 &64.0 &68.9 \\
MMT \cite{ge2020mutual} & ICLR 2020 & 22.9 & 49.2 & 63.1 & 68.8 & 23.3 & 50.1 & 63.9 & 69.8 \\
JVTC+ \cite{li2020joint} &ECCV 2020 &25.1 &48.6 &65.3 &68.2 &27.5 &52.9 &\underline{70.5} &\underline{75.9} \\
SpCL \cite{ge2020self} &NeurIPS 2020 &\underline{26.8} &53.7 &65.0 &69.8 &26.5 &53.1 &65.8 &70.5 \\
Dual-Refinement \cite{dai2020dual} &TIP 2021 &25.1 &53.3 &66.1 &71.5 &26.9 &55.0 &68.4 &73.2 \\
UNRN \cite{zheng2020exploiting} & AAAI 2021 & 25.3 & 52.4 & 64.7 &69.7 & 26.2 & 54.9 & 67.3 & 70.6 \\ 
GLT \cite{zheng2021group} &CVPR 2021 &26.5  &\underline{56.6}  &\underline{67.5}  &\underline{72.0}  &\underline{27.7}  &\underline{59.5}  &70.1  &74.2  \\ \hline
IDM (Ours) & ICCV 2021 &\textbf{33.5} &\textbf{61.3} &\textbf{73.9} &\textbf{78.4} &\textbf{35.4} &\textbf{63.6} &\textbf{75.5} &\textbf{80.2} \\ \hline
\end{tabular}
\end{center}
\vspace{-1.5em}
\end{table*}

\begin{table*}[tp]
\caption{Comparison with the state-of-the-art  UDA re-ID methods on synthetic $\rightarrow$ real tasks.
%\protect\footnotemark[1]
}
\label{tab:SOTA-synthetic2real}
\footnotesize
\begin{center}
\begin{tabular}{l|c|cccc|cccc|cccc}
\hline
\multirow{2}{*}{Methods} & \multirow{2}{*}{Reference} & \multicolumn{4}{c|}{PersonX $\to$ Market-1501} & \multicolumn{4}{c|}{PersonX $\to$ DukeMTMC-reID} & \multicolumn{4}{c}{PersonX $\to$ MSMT17} \\ \cline{3-14} 
 &  & mAP & R1 & R5 & R10 & mAP & R1 & R5 & R10 & mAP & R1 & R5 & R10 \\ \hline
MMT \cite{ge2020mutual} & ICLR 2020 &71.0 &86.5 &94.8 &\underline{97.0} &60.1 &74.3 &86.5 &90.5  &17.7  &39.1  &52.6  &58.5  \\
SpCL \cite{ge2020self} & NeurIPS 2020 &\underline{73.8} &\underline{88.0} &\underline{95.3} &96.9 &\underline{67.2} &\underline{81.8} &\underline{90.2} &\underline{92.6} &\underline{22.7} &\underline{47.7} &\underline{60.0} &\underline{65.5}  \\
IDM (Ours) & ICCV 2021 &\textbf{81.3} &\textbf{92.0} &\textbf{97.4} &\textbf{98.2} &\textbf{68.5} &\textbf{82.6} &\textbf{91.2} &\textbf{93.4} &\textbf{30.3} &\textbf{58.4} &\textbf{70.7} &\textbf{75.5}  \\ \hline \hline
\multirow{2}{*}{Methods} & \multirow{2}{*}{Reference} & \multicolumn{4}{c|}{Unreal $\to$ Market-1501} & \multicolumn{4}{c|}{Unreal $\to$ DukeMTMC-reID} & \multicolumn{4}{c}{Unreal $\to$ MSMT17} \\ \cline{3-14} 
 &  & mAP & R1 & R5 & R10 & mAP & R1 & R5 & R10 & mAP & R1 & R5 & R10 \\ \hline
JVTC \cite{li2020joint} & ECCV 2020 &78.3 &90.8 &- &- &66.1 &81.2 &- &- &25.0 &53.7 &- &-  \\
IDM (Ours) &ICCV 2021 &\textbf{83.2} &\textbf{92.8} &\textbf{97.3} &\textbf{98.2} &\textbf{72.4} &\textbf{84.6} &\textbf{92.0} &\textbf{94.0} &\textbf{38.3} &\textbf{67.3} &\textbf{78.4} &\textbf{82.6} \\ \hline
\end{tabular}
\end{center}
\vspace{-2em}
\end{table*}

\textbf{Comparisons on real $\to$ real UDA re-ID tasks.}
The existing UDA re-ID methods evaluated on real $\to$ real UDA tasks can be mainly divided into three categories based on their training schemes. (1) GAN transferring methods include PTGAN \cite{wei2018person}, SPGAN+LMP \cite{deng2018image}, \cite{zhong2018generalizing}, and PDA-Net \cite{li2019cross}. 
(2) Fine-tuning methods include PUL \cite{fan2018unsupervised}, PCB-PAST \cite{zhang2019self}, SSG \cite{fu2019self}, AD-Cluster \cite{zhai2020ad}, MMT \cite{ge2020mutual}, NRMT \cite{zhao2020unsupervised}, MEB-Net \cite{zhai2020multiple}, Dual-Refinement \cite{dai2020dual}, UNRN \cite{zheng2020exploiting}, and GLT \cite{zheng2021group}. 
(3) Joint training methods commonly use the memory bank \cite{xiao2017joint}, including ECN \cite{zhong2019invariance}, MMCL \cite{wang2020unsupervised}, ECN-GPP \cite{zhong2020learning}, JVTC+ \cite{li2020joint}, and SpCL \cite{ge2020self}. However, all these methods neglect the significance of intermediate domains, which can smoothly bridge the domain adaptation between the source and target domains to better transfer the source knowledge to the target domain. Instead, we propose an IDM module to generate the appropriate intermediate domains to better improve the performance of UDA re-ID. As shown in Table \ref{tab:SOTA-real2real}, our method can outperform the second best UDA re-ID methods by a large margin on all these benchmarks.

\textbf{Comparisons on synthetic $\to$ real UDA re-ID tasks.} Compared with the real $\to$ real UDA re-ID tasks, the synthetic $\to$ real UDA tasks are more challenging because the domain gap between the synthetic and real images are often larger than that between the real and real images. As shown in Table \ref{tab:SOTA-synthetic2real}, our method can outperform the SOTA methods by a large margin where mAP of our method is higher than SpCL~\cite{ge2020self} by 7.5\%, 1.3\%, and 7.6\% when testing on Market-1501, DukeMTMC-reID, and MSMT17 respectively. All these significant results of our method have shown the superiority of our proposed IDM module that can generate the appropriate intermediate domains to better improve the performance on UDA re-ID.

\section{Conclusion}
This paper proposes a plug-and-play Intermediate Domain Module (IDM) to tackle the problem of UDA re-ID. From a novel perspective that intermediate domains can bridge the source and target domains, our purposed IDM module can generate appropriate intermediate domain representations to better transfer the source knowledge to improve the model's discriminability on the target domain. The intermediate domains' distribution is controlled by the two domain factors generated by the IDM module. Specifically, we propose the bridge losses to enforce the intermediate domains to be located onto the appropriate path between the source and target domains in a manifold. Besides, we also propose a diversity loss to constrain the domain factors to prevent the intermediate domains from  being over-fitting to either of the source and target domains. Extensive experiments have shown the effectiveness of our method.

\noindent \textbf{Acknowledgements:} This work was supported by the National Natural Science Foundation of China under Grant 62088102, and in part by the PKU-NTU Joint Research Institute (JRI) sponsored by a donation from the Ng Teng Fong Charitable Foundation, and was partially supported by SUTD Project PIE-SGP-Al2020-02. 

{
\balance
\small
\bibliographystyle{ieee_fullname}
\bibliography{egbib}
}

\clearpage

{\section*{\Large Appendix}}
\setcounter{equation}{0}
\setcounter{subsection}{0}
\setcounter{section}{0}
\renewcommand{\theequation}{A.\arabic{equation}}
\renewcommand\thesection{\Alph{section}}

In this supplementary material, we provide more details that could not be presented in the regular paper due to the space limitation. In Section \ref{sec:train_test}, we show the overall training and testing procedure. In Section \ref{sec:implement}, we provide more details about the implementation of our method. 
In Section~\ref{sec:additional_experiments}, we compare our method with the state-of-the-arts on other two  real $\to$ real UDA re-ID benchmarks. 
In Section~\ref{sec:parameters}, we analyse the effects of the hyper-parameters.

\section{Overall Training and Testing Procedure}
\label{sec:train_test}
The overall training procedure of our method is shown in Algorithm \ref{alg:algorithm1}, where we use XBM \cite{wang2020cross} as the memory bank to implement our Strong Baseline method. If using XBM, it means we implement our proposed IDM module in Strong Baseline; If not using XBM, it means we implement the IDM module in Naive Baseline. More details about XBM can be seen at Section \ref{sec:xbm} below.
Our proposed IDM module is only used for training and is discarded for testing. In the testing procedure, we use the L2-normalized features after the global average pooling (GAP) layer followed by a batch normalization (BN) layer.

\section{Implementation Details}
\label{sec:implement}

ResNet-50 \cite{he2016deep} pretrained on ImageNet is adopted as the backbone network. Domain-specific BNs \cite{chang2019domain} are used in the backbone network to narrow the domain gaps.
Following \cite{luo2019strong}, we resize the image size to 256$\times$128 and apply some common image augmentation techniques, including random flipping, random cropping, and random erasing \cite{zhong2020random}. We perform DBSCAN \cite{ester1996density} clustering on the unlabeled target data to assign pseudo labels at the beginning of each training epoch, in a manner like the existing UDA re-ID methods \cite{fu2019self,song2020unsupervised,ge2020self}.
The mini-batch size is 128, including 64 source images of 16 identities and 64 target images of 16 pseudo identities. We totally train 50 epochs and each epoch contains 400 iterations. The initial learning rate is set as $3.5\times 10^{-4}$ which will be divided by 10 at the 20th and 40th epoch respectively. The Adam optimizer with weight decay $5\times 10^{-4}$ and momentum 0.9 is adopted in our training. 
The loss weights $\mu_{1}, \mu_{2}, \mu_{3}$ are set as 0.7, 0.1, 1 respectively. 
In the IDM module, the FC1 layer is parameterized by $W_{1}\in \mathbb{R}^{c\times 2c}$  and MLP is composed of two fully connected layers which are parameterized by $W_{2}\in \mathbb{R}^{(c/r) \times c}$ and $W_{3}\in \mathbb{R}^{2 \times (c/r)}$ respectively, where $c$ is the representations' channel number after the $m$-th stage and $r$ is the reduction ratio. If not specified, we plug the IDM module after the stage-0 of ResNet-50 and set $r$ as 2. The IDM module is only used in training and will be discarded in testing. Our method is implemented with Pytorch, and four NVIDIA RTX 2080Ti GPUs are used for training and only one GPU is used for testing.

\subsection{Clustering on the Target Domain}
As shown in Algorithm \ref{alg:algorithm1}, we perform DBSCAN \cite{ester1996density} clustering on the features of all the target domain samples to assign pseudo labels for the target domain samples, which is similar to the the existing clustering-based UDA re-ID methods \cite{fu2019self,song2020unsupervised}. Specifically, we use the the Jaccard distance~\cite{zhong2017re} as the metric in DBSCAN, where the $k$-reciprocal nearest neighbor set is used to calculate the pair-wise similarity. We set $k$ as 30 in our experiments. In DBSCAN, we set the maximum distance between neighbors as 0.6 and the minimal number of neighbors for a dense point as 4.

\begin{algorithm}[htp]
\small 
%\footnotesize
\caption{The overall training procedure}
\label{alg:algorithm1}
\KwIn{
Source labeled dataset $\{(x^{s}_{i},y^{s}_{i})\}$ and target unlabeled dataset $\{x^{t}_{i}\}$;
}
\KwOut{
The trained backbone network $f(\cdot)$ and classifier $\varphi(\cdot)$;
}
    
Initialize the backbone network $f(\cdot)$ with the ImageNet-pretrained ResNet-50;

Initialize the XBM memory as an empty queue $M$;

Plug our IDM module after the $m$-th stage of the backbone $f(\cdot)$ and randomly initialize it.

\For{$epoch=1$ to MaxEpochs}
{
Use the backbone $f(\cdot)$ to extract features $\{f^{t}_{i}\}$ for the target dataset $\{x^{t}_{i}\}$;

Assign pseudo labels $\{y^{t}_{i}\}$ for target domain samples $\{x^{t}_{i}\}$ by performing DBSCAN clustering on the target features $\{f^{t}_{i}\}$;

\For{$iter=1$ to MaxIters}
{

Sample a mini-batch of samples including $n$ source samples $\{(x^{s}_{i},y^{s}_{i})\}_{i=1}^{n}$ and $n$ target samples $\{(x^{t}_{i},y^{t}_{i})\}_{i=1}^{n}$;

Feed forward the batch into the network to obtain the features and predictions for the source, target, and intermediate domains: $(f^{s},\varphi^{s})$, $(f^{t},\varphi^{t})$, and $(f^{\rm inter},\varphi^{\rm inter})$;

\If {using XBM}{

Enqueue($M$, $\{(f^{t},y^{t})\}$, $\{(f^{s},y^{s})\}$);

\If {M is full}{
Dequeue($M$);
}
Use all entries in $M$ for the hard negatives mining in the triplet loss in $\mathcal L_{\rm ReID}$ from Eq.~(\ref{eq:overall_loss});

}

Calculate the overall training loss by Eq. (\ref{eq:overall_loss});

Update $f(\cdot)$, $\varphi(\cdot)$, and our IDM module together by back-propagating the gradients of Eq. (\ref{eq:overall_loss});
}
}
\end{algorithm}
%\vspace{-1em}

\begin{table*}[htp]
\caption{Comparison with the state-of-the-art UDA re-ID methods on other real $\to$ real tasks.
}
\small
\label{tab:SOTA_real2real}
\begin{center}
\begin{tabular}{l|c|p{1cm}<{\centering}p{1cm}<{\centering}p{1cm}<{\centering}p{1cm}<{\centering}|p{1cm}<{\centering}p{1cm}<{\centering}p{1cm}<{\centering}p{1cm}<{\centering}}
\hline
\multirow{2}{*}{Methods} & \multirow{2}{*}{Reference} & \multicolumn{4}{c|}{MSMT17 $\to$ Market-1501} & \multicolumn{4}{c}{MSMT17 $\to$ DukeMTMC-reID} \\ \cline{3-10} 
 &  & mAP & R1 & R5 & R10 & mAP & R1 & R5 & R10 \\ \hline
CASCL \cite{wu2019unsupervised} &ICCV 2019 &35.5 &65.4 &80.6 &86.2 &37.8 &59.3 &73.2 &77.8 \\
MAR \cite{yu2019unsupervised} &CVPR 2019 &40.0 &67.7 &81.9 &87.3 &48.0 &67.1 &79.8 &84.2 \\ 
PAUL \cite{yang2019patch} &CVPR 2019 &40.1 &68.5 &82.4 &87.4 &53.2 &72.0 &82.7 &86.0 \\ 
DG-Net++ \cite{zou2020joint} &ECCV 2020 &64.6 &83.1 &91.5 &94.3 &58.2 &75.2 &73.6 &86.9 \\  
D-MMD \cite{mekhazni2020unsupervised} &ECCV 2020 &50.8 &72.8 &88.1 &92.3 &51.6 &68.8 &82.6 &87.1 \\ 
MMT-dbscan \cite{ge2020mutual} &ICLR 2020 &75.6 &89.3 &95.8 &97.5 &63.3 &77.4 &88.4 &91.7 \\ 
SpCL \cite{ge2020self} &NeurIPS 2020 &\underline{77.5} &\underline{89.7} &\underline{96.1} &\underline{97.6} &\underline{69.3} &\underline{82.9} &\underline{91.0} &\underline{93.0} \\ \hline
IDM (Ours) & ICCV 2021 &\textbf{82.1} &\textbf{92.4} &\textbf{97.5} &\textbf{98.4} &\textbf{71.9} &\textbf{83.6} &\textbf{91.5} &\textbf{93.4} \\ \hline
\end{tabular}
\end{center}
\vspace{-1.5em}
\end{table*}

\subsection{The structure of our IDM module}
Our proposed IDM module is very easy to implement, including a FC1 layer and a MLP
(Muti-Layer Perception) followed by a softmax function. Specifically, the FC1 layer is a fully connected layer parameterized by $W_{1}\in \mathbb{R}^{c\times 2c}$. The MLP contains two fully connected layers which are parameterized by $W_{2}\in \mathbb{R}^{(c/r)\times c}$ and $W_{3}\in \mathbb{R}^{2\times(c/r)}$ respectively. We denote $c$ as the channel number of the feature map at the $m$-th stage of ResNet-50, and denote $r$ as the reduction ratio for the dimension reduction. In ResNet-50, the channel number $c$ is 64/256/512/1024/2048 after the stage-0/1/2/3/4 respectively.
In Figure \ref{fig:reduction-ratio}, we evaluate the effectiveness of different values on the reduction ratio $r$.

\subsection{Implementation of XBM}
\label{sec:xbm}
Similar to many joint training UAD re-ID methods \cite{zhong2019invariance,zhong2020learning,wang2020unsupervised,li2020joint,ge2020self,zheng2020exploiting} that use the memory bank \cite{xiao2017joint,wu2018unsupervised,wang2020cross} to improve the discriminability on the target domain, we also use the memory bank to implement our Strong Baseline method. Specifically, we use the memory bank to mine hard negatives of both source and target domains to calculate the triplet loss, similarly to XBM~\cite{wang2020cross}. Following XBM \cite{wang2020cross}, the memory bank is maintained and updated as a queue: for each mini-batch, we enqueue the features and (pseudo) labels of samples in this current mini-batch, and dequeue the entites of the earlist mini-batch if the queue is full. 
For each mini-batch, we use all the entites in the memory bank to mine hard negatives for the triplet loss \cite{hermans2017defense}.
The procedure of maintaining and updating the XBM can be seen in Algorithm~\ref{alg:algorithm1}. 
Besides, we set the memory ratio as $R_{M}=N^{M}/(N^{s}+N^{t})$, where $N^{M}$ is the size the memory bank and $N^{s}$ ($N^{t}$) is the number of all the source (target) training samples. If not specified, we set $R_{M}$ as 1 in our experiments, \textit{i.e.,} the size of the memory is the same as the size of the whole training dataset (including both source and target domains). We also evaluate the effectiveness on different values of $R_{M}$ in Figure \ref{fig:memory-ratio}.

\section{Additional Experimental Results}
\label{sec:additional_experiments}
Some other real $\to$ real tasks are used to evaluate the UDA re-ID performances in the existing UDA re-ID methods \cite{wu2019unsupervised,yu2019unsupervised,yang2019patch,zou2020joint,mekhazni2020unsupervised,ge2020self}, where they use MSMT17~\cite{wei2018person} as the source dataset, and use Market-1501 \cite{zheng2015scalable} and DukeMTMC-reID \cite{ristani2016performance,zheng2017unlabeled} as the target datasets respectively. As shown in Table \ref{tab:SOTA_real2real}, our method can outperform the state-of-the-arts methods on these two real $\to$ real tasks by a large margin. From all the results in our regular paper and this supplementary material, our method can significantly outperform the state-of-the-arts methods in all the existing UDA re-ID benchmarks.

\section{Parameter Analysis}
\label{sec:parameters}
We tune the hyper-parameters on the task of Market $\to$ Duke, and apply the tuned hyper-parameters to all the other UDA re-ID tasks in our regular paper.

\begin{figure}[tp]
    \centering
    \includegraphics[width=\linewidth]{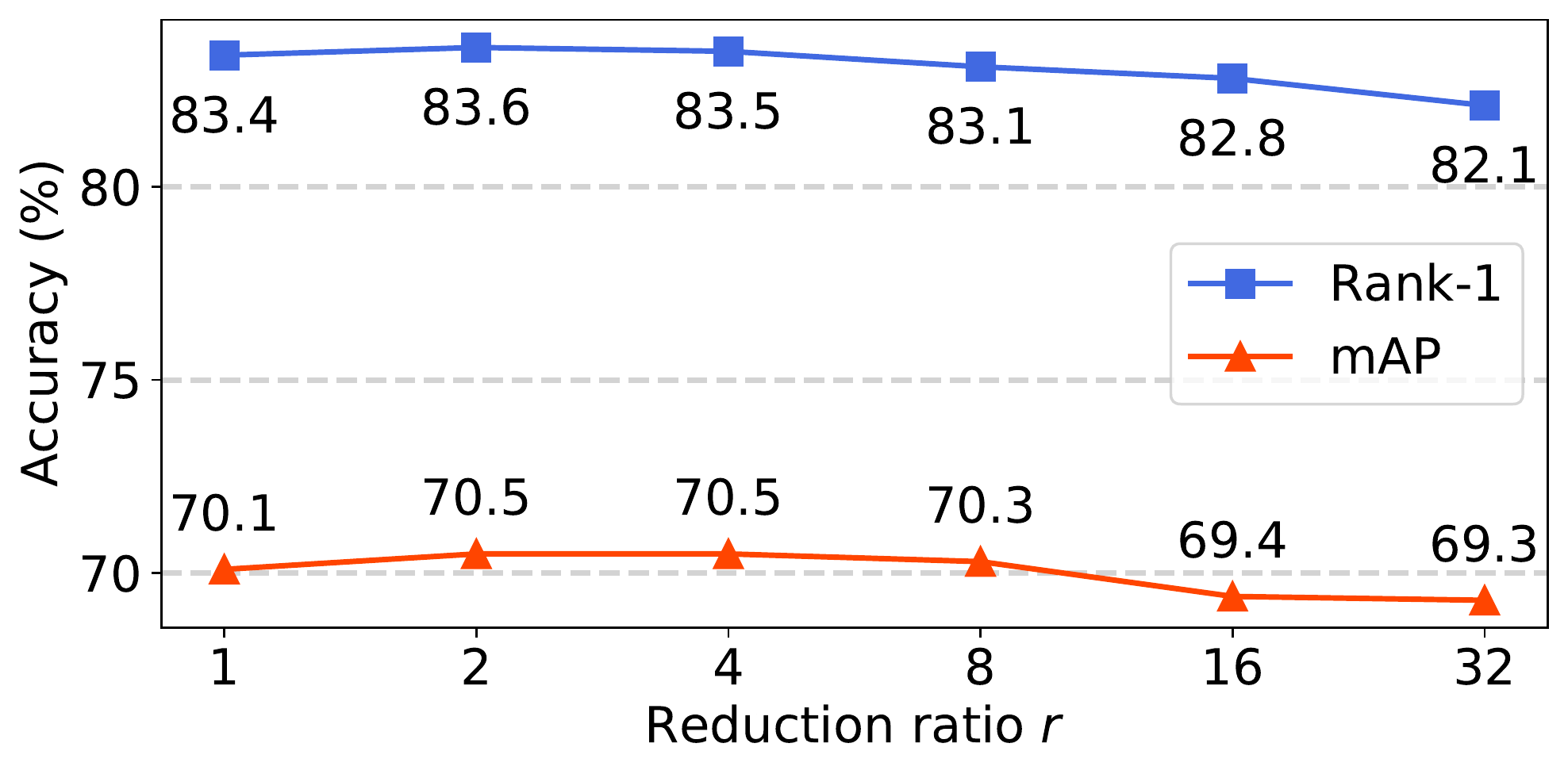}
    \caption{Performance of our method with different values of the reduction ratio $r$ in our IDM module. Evaluating on Market $\to$ Duke when our IDM is plugged after the stage-0 of ResNet-50.}
    \label{fig:reduction-ratio}
\end{figure}

\begin{figure}[tp]
    \centering
    \includegraphics[width=\linewidth]{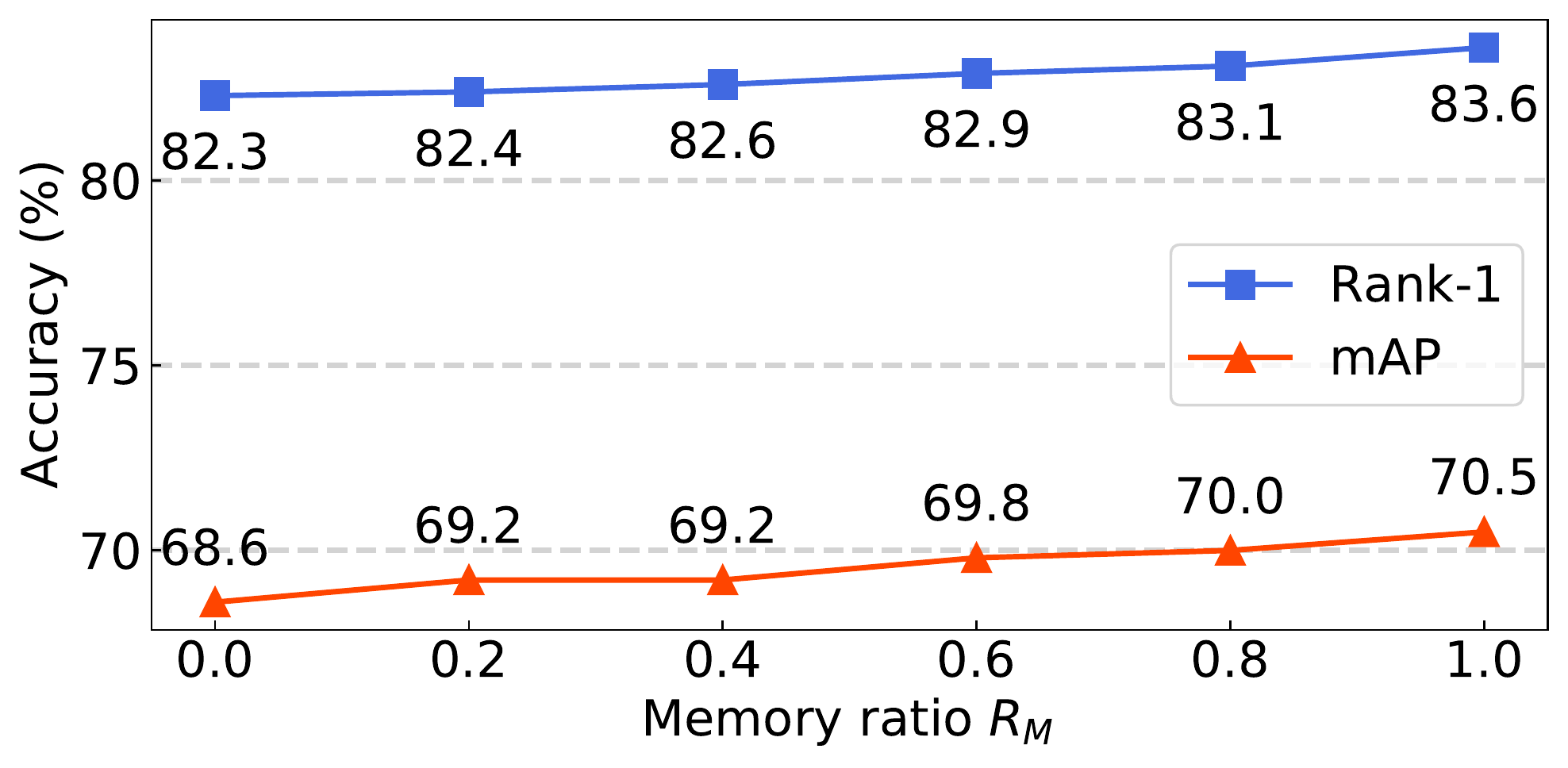}
    \caption{Performance of our method with different values of the memory ratio  $R_{M}$ in XBM.}
    \label{fig:memory-ratio}
\end{figure}

\begin{figure}[htp]
    \centering
    \includegraphics[width=\linewidth]{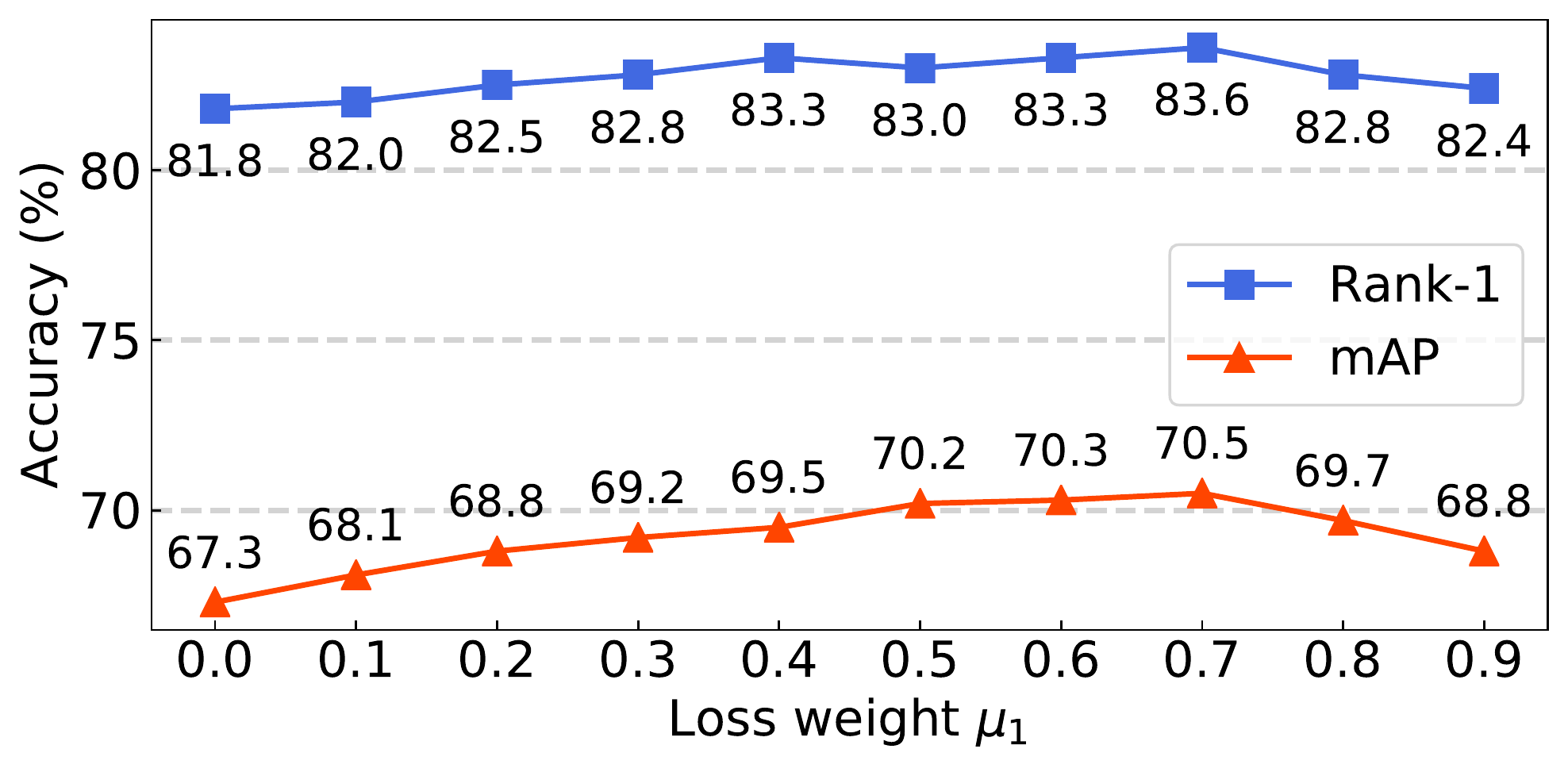}
    \caption{Performance of our method with different values of the loss weight  $\mu_{1}$.}
    \label{fig:mu1}
    %\vspace{-0.5em}
\end{figure}

\begin{figure}[htp]
    \centering
    \includegraphics[width=\linewidth]{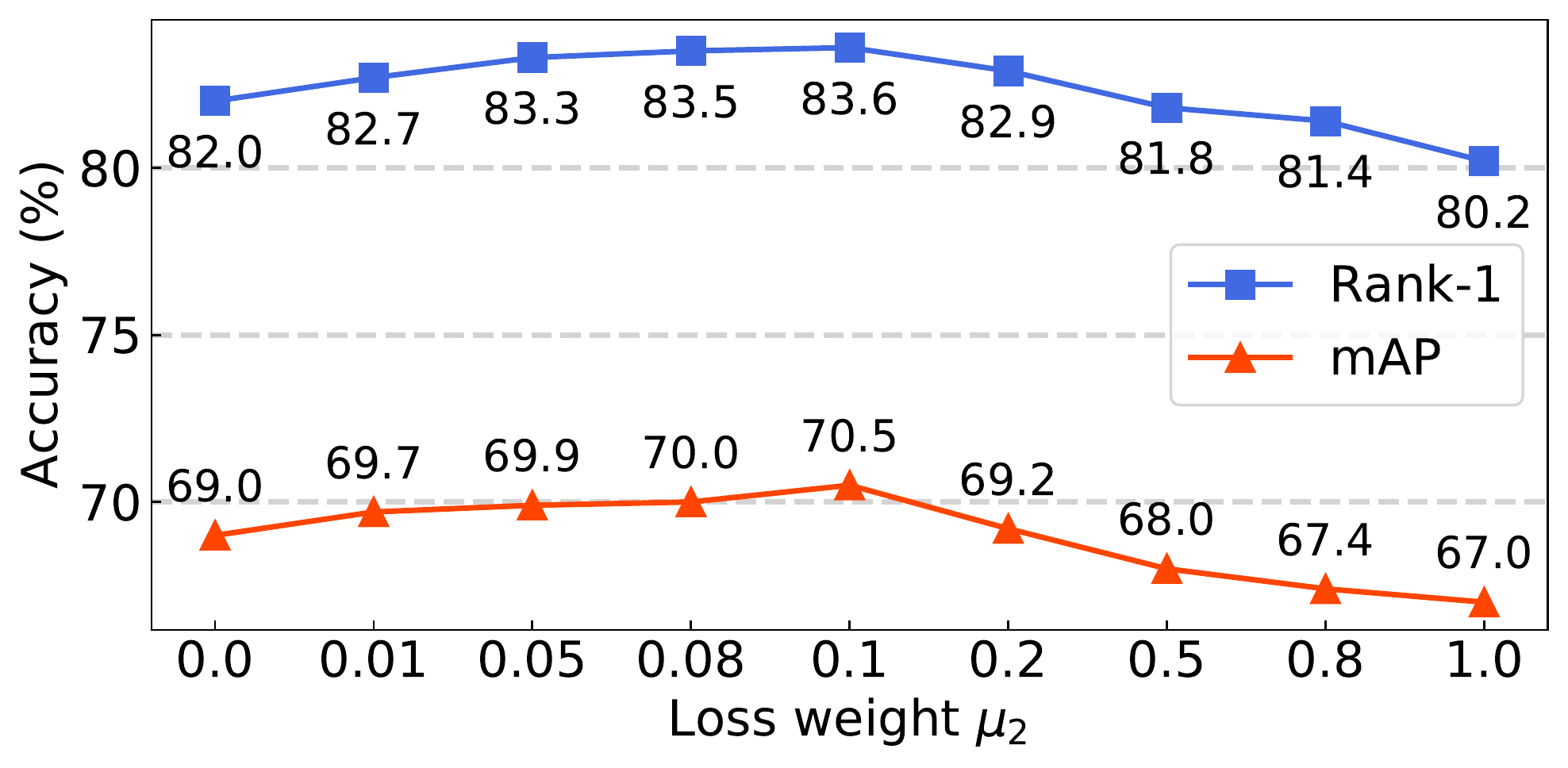}
    \caption{Performance of our method with different values of the loss weight  $\mu_{2}$.}
    \label{fig:mu2}
\end{figure}

\begin{figure}[htp]
    \centering
    \includegraphics[width=\linewidth]{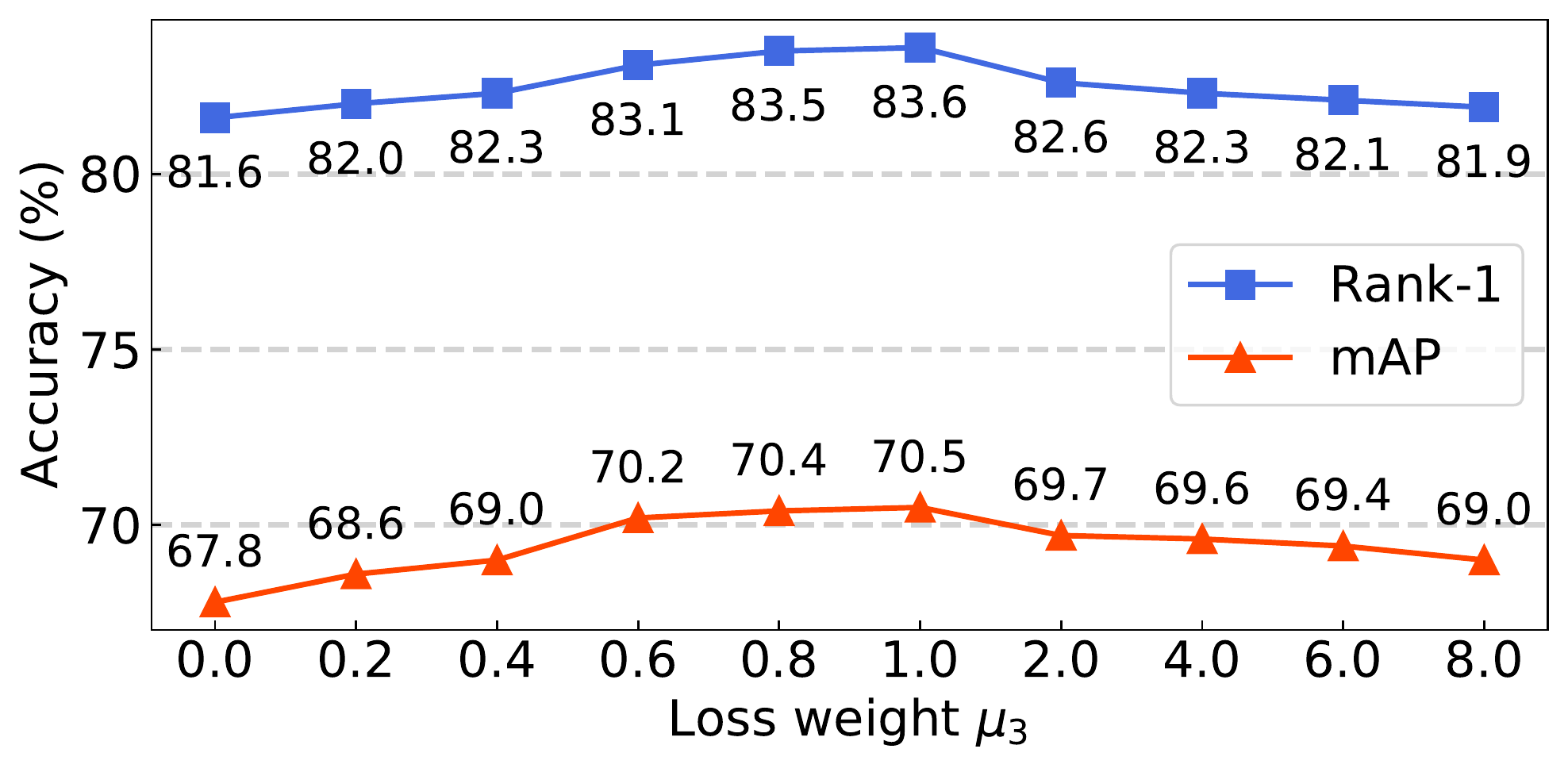}
    \caption{Performance of our method with different values of the loss weight  $\mu_{3}$.}
    \label{fig:mu3}
    %\vspace{-0.5em}
\end{figure}

\subsection{The reduction ratio $r$ in our IDM module}
In Figure \ref{fig:reduction-ratio}, we evaluate the effectiveness of different values on the reduction ratio $r$ when we plug our IDM module after the stage-0 of ResNet-50. When $r$ gets larger, the performance gets slightly lower because the larger reduction ratio will make the IDM harder to learn. From the results in Figure \ref{fig:reduction-ratio}, we set $r$ as 2 for our method in all the other UDA re-ID tasks.

\subsection{The memory ratio $R_{M}$ for the XBM in our Strong Baseline.}
We implement XBM \cite{wang2020cross} in our Strong Baseline, where the memory bank is set as a queue of the size $N^{M}$. We use the memory ratio $R_{M}=N^{M}/(N^{s}+N^{t})$ to control the size of the memory bank, where $N^{s}$ ($N^{t}$) is the size of the source (target) domain training dataset. We evaluate the performance on Market $\to$ Duke in Figure \ref{fig:memory-ratio}. When $R_{M}=0$, it means we implement our method based on Naive Baseline, \textit{i.e.,} ``Baseline1 + Our IDM (full)'' in Table \ref{tab:ablation} in our paper submission. When $R_{M}=1$, it means we implement our method based on Strong Baseline, \textit{i.e.,} ``Baseline2 + Our IDM (full)'' in Table \ref{tab:ablation} in our paper submission. When the memory size gets larger, the performance will get better because the larger memory bank can mine more effective negatives for the target domain. However, whether we use XBM or not, our method can outperform the baseline method by a large margin.

\subsection{The loss weight $\mu_{1}$}
We tune the value of the loss weight $\mu_{1}$ in Figure \ref{fig:mu1}, where $\mu_{1}$ is the weight to balance the bridge loss $\mathcal L_{\rm bridge}^{\varphi}$ in Eq. (\ref{eq:overall_loss}) in our regular paper. When $\mu_{1}=0$, it means ``Baseline2 + Our IDM w/o $\mathcal L_{\rm bridge}^{\varphi}$''. We use this bridge loss $\mathcal L_{\rm bridge}^{\varphi}$ to enforce on intermediate domains' prediction space. As shown in Figure \ref{fig:mu1}, the performance gets better when $\mu_{1}$ ranges from 0 to 0.7. If $\mu$ varies from 0.7 to 0.9, the performance will get a little degradation because more penalization on intermediate domains' prediction space will affect the learning of the source and target domains. Thus, we set $\mu_{1}$ as 0.7 in all the other experiments in our regular paper.

\subsection{The loss weight $\mu_{2}$}
We compare the performance of different values of the loss weight $\mu_{2}$ in Figure \ref{fig:mu2}. 
The weight $\mu_{2}$ is used to balance the bridge loss $\mathcal L_{\rm bridge}^{f}$ in Eq. (\ref{eq:overall_loss}). We use $\mathcal L_{\rm bridge}^{f}$ to enforce on intermediate domains' feature space to keep the right distance between intermediate domains to the source and target domains. When $\mu_{2}=0$, it is the same as ``Baseline2 + Our IDM w/o $\mathcal L_{\rm bridge}^{f}$'' in Table \ref{tab:ablation} in our regular paper. When $\mu_{2}$ gets close to 0.1, the performance gets better. If setting a large weight value of $\mu_{2}$, it will bring a little performance degradation because the overall loss will penalize more on the intermediate domains' feature space while penalize less on the source and target domains. From Figure~\ref{fig:mu2}, we set $\mu_{2}$ as an appropriate value 0.1 to better balance the bridge loss $\mathcal L_{\rm bridge}^{f}$ in Eq. (\ref{eq:overall_loss}).

\subsection{The loss weight $\mu_{3}$}
In Figure \ref{fig:mu3}, we evaluate the performances of our method with different values of the loss weight $\mu_{3}$. We use the weight $\mu_{3}$ to balance the diversity loss $\mathcal L_{\rm div}$ in Eq. (\ref{eq:overall_loss}). When $\mu=0$, it means the performance of ``Baseline2 + Our IDM w/o $\mathcal L_{\rm div}$'' in Table~\ref{tab:ablation} in our regular paper. As the results reported in Figure \ref{fig:mu3}, we set $\mu_{3}$ as 1.0 for the experiments in all the other UDA re-ID tasks.

\end{document}